\documentclass[]{fairmeta}

\title{ADEPTS: A Capability Framework for \\ Human-Centered Agent Design}

\author[1]{Pierluca D'Oro}
\author[1]{Caley Drooff}
\author[1]{Joy Chen}
\author[1]{Joseph Tighe}

\affiliation[1]{FAIR at Meta}

\abstract{
Large language models have paved the way to powerful and flexible AI agents, assisting humans by increasingly integrating into their daily life. This flexibility, potential, and growing adoption demands a holistic and cross-disciplinary approach to developing, monitoring and discussing the capabilities required for agent-driven user experiences. However, current guidance on human-centered AI agent development is scattered: UX heuristics focus on interface behaviors, engineering taxonomies describe internal pipelines, and ethics checklists address high-level governance. There is no concise, user-facing vocabulary that tells teams what an agent should fundamentally be able to do. We introduce ADEPTS, a capability framework defining a set of core user-facing capabilities to provide unified guidance around the development of AI agents. ADEPTS is based on six principles for human-centered agent design, that express the minimal, user-facing capabilities an AI agent should demonstrate to be understandable, controllable and trustworthy in everyday use. ADEPTS complements existing frameworks and taxonomies; differently from them, it sits at the interface between technical and experience development. By presenting ADEPTS, we aim to condense complex AI-UX requirements into a compact framework that is actionable guidance for AI researchers, designers, engineers, and policy reviewers alike. We believe ADEPTS has the potential of accelerating the improvement of user-relevant agent capabilities, of easing the design of experiences that take advantage of those capabilities, and of providing a shared language to track and discuss progress around the development of AI agents.
}

\date{\today}
\correspondence{Pierluca D'Oro at \email{pierluca@meta.com}}

\begin{document}

\maketitle
\tcbset{
  enhanced,
  frame hidden,
  left=0.5cm,
  right=0.5cm,
  top=0.5cm,
  bottom=0.5cm,
  arc=10pt,
  colback=metabg,
  before skip=8pt, 
  after skip=8pt,
}

\tableofcontents

\section{Introduction}
\label{section:intro}
AI agents autonomously perceive the world and act upon it to execute actions towards a specified objective~\citep{russell2016artificial,sutton1998reinforcement}.
Large Language Models (LLMs) revolutionized the way agents are developed, drastically accelerating their adoption~\citep{brown2020language,ouyang2022training}.
Since LLMs empower AI agents with the ability to use natural language proficiently, alongside with extensive knowledge, expectations around modern AI agents are evolving well-beyond mere task execution.
AI agents are asked to possess capabilities that can smoothly power \emph{user experiences}.
These capabilities should allow AI agents not only to autonomously execute tasks on behalf of humans, but also to offer the technology to create usable systems.

In this evolving world, as the adoption of AI agents into people's daily life grows, there is also a growing need for a shared language to measure progress in their development and to accelerate it along the axes that are maximally beneficial for creating empowering, pleasurable, and accessible experiences for humans.
This shared language should facilitate \emph{human-centered agent design}: a paradigm to build agents that directly considers the way their capabilities can be beneficial to humans~\citep{ozmen2023six,russell2022human}.  
Specifically, there is a need for a conceptual framework that can be used by researchers, engineers, designers, cross-discipline product teams, and by the public, to understand and guide the evolution of AI agent's capabilities relative to improving user experiences.

Existing frameworks do not offer a capability-focused language that can be shared cross-functionally to inform agent development.
Instead, they either focus on UX-specialized heuristics and considerations, not directly concerned around an agent's capabilities~\citep{amershi2019guidelines,google_pair_guidebook_2025}, or on engineering and research considerations, not necessarily surfaced to humans~\citep{yehudai2025survey,build_agents_anthropic}.

In this paper, we propose ADEPTS, a capability framework for human-centered agent design.
To develop ADEPTS, we start from a set of principles for human-centered agent design.
We propose six of these principles, and craft ADEPTS as a concise six-point set of capabilities that an agent should possess to power modern user experiences.
ADEPTS takes its name from the six capabilities that we identified: \textbf{A}ctuation, \textbf{D}isambiguation, \textbf{E}valuation, \textbf{P}ersonalization, \textbf{T}ransparency, and \textbf{S}afety.
An AI agent that performs well according to these capabilities will contribute to creating systems that are both low-friction (mainly via actuation, disambiguation and personalization) and trustworthy (mainly via evaluation, transparency and safety), two fundamental features required for creating high-quality user experiences~\citep{nielsen1995conduct}.

By construction, ADEPTS focuses on \emph{what} an agent should be able to do to support everyday experiences for users, clustering user-facing concerns into a \emph{capability contract} that an agent should satisfy.
ADEPTS does not prescribe any \emph{how}, both in terms of user interface and technical approach, leaving space for product and agent designers to tailor experience and model development to their specific needs.
In other words, working at the \emph{interaction layer}, our framework describes how a human should experience an agent, not the details of how to build the agent or how it should exactly work.
ADEPTS is granular enough to guide agent design, but broad enough to guide UX design or even policy: it can be used to map UX patterns against the capabilities, or to measure readiness of a given AI agent to power a usable experience.

To build additional common ground to discuss and accelerate the development of AI agents, we go beyond the list of high-level capabilities, and provide, for each one of the capabilities, reference tiers to assess the level of competency of AI agents along different axes.
A number of these reference tiers are derived, and brought together for a comprehensive view, from existing capability frameworks.
We hope that these reference tiers can serve both as references for high-level discussions, but especially for inspiring the development of application-specific benchmarks to measure the ADEPTS capabilities.

We designed ADEPTS to be general enough to help formalizing the design of various types of agents.
Although it has some inherent limitations, the capability contract ADEPTS prescribes is generally valid on a wide class of underlying technologies or embodiments that an AI agent can use, offering a common platform for cross-embodiment and cross-application capability development.

\section{Principles for Human-Centered Agent Design}
We believe that, to help the development of AI agents that can power a wide range of user experiences, it can be beneficial to refer to an evaluation framework that is directly focused on human-centered agent design.
At a fundamental level, this framework should answer the question: \emph{what should an AI agent be capable of doing in order to power usable and trustworthy systems?}

To answer this question, we propose six principles for human-centered agent design, which aim to capture the capabilities that an AI agent should possess to support a range of user experiences.

\begin{tcolorbox}
\textbf{Principle 1: Autonomous Actuation} \\
The agent should autonomously execute tasks on the user's behalf, translating intent into actions according to the permissions granted by the user and the constraints enforced by the agent designer.
\end{tcolorbox}

The first principle is framed around the defining quality of AI agents, autonomously executing actions in an environment for a user.
Crucially, the principle puts focus on the fact that not only an agent should be able to execute a task, but that this also implies being able to understand what the user intent is, before translating it into actions.
Moreover, the principle also highlights that an agent should not violate permissions, communicated by the user via their instructions, and constraints, enforced by an agent designer by e.g., choosing an appropriate action space~\citep{pmlr-v270-esser25a}.

\begin{tcolorbox}
\textbf{Principle 2: Intent Disambiguation} \\
The agent should actively clarify and confirm the user's goal, context, and constraints whenever uncertainty could alter the outcome.
\end{tcolorbox}

The second principle focuses on disambiguation of the intent of a user. 
It is a natural expectation for modern AI agents to be able to be \emph{promptable}, i.e., to support receiving open-ended instruction formats (e.g., via natural language~\citep{wang2024survey}).
Thus, while instructing the agent to achieve a certain goal, there might be inherent ambiguity in the way a user communicates it to the agent.
This can happen because of a user's unawareness of the need of specifying a task that is non-ambiguous, a user's lack of knowledge about the functionalities and constraints of the agent's embodiment, or even just due to the ambiguity of natural language as a mean of communication~\citep{chomsky2002syntactic}. 

Following the idea of human-centered agent design, an agent designer cannot make the assumption that a user will always specify a complete and unambiguous goal to the system; instead, the AI agent should have the capability to reactively and proactively disambiguate, making the user aware of the system's constraints and interacting with them to understand their intent. 

\begin{tcolorbox}
\textbf{Principle 3: Situational Evaluation} \\
The agent should track task progress and overall context, surfacing status and providing answers for the user to understand the current situation or resume control.
\end{tcolorbox}

The third principle revolves around the need for the agent to have the capability to evaluate the situation it is operating in, and to interact with the user to surface it.
This capability has a crucial role for an agent to power a usable system: being able to detect success could be crucial for both error handling and for increasing user trust.
Moreover, it is reasonable to think that most users interacting with an AI agent would expect an agent that is able to execute actions for them to also be able to proficiently answer questions about the world they are operating in, or understand the consequences of certain actions.
Situational evaluation is also paramount for appropriately requesting human intervention, which is key to the dynamics of most collaborative human-AI user experiences.

\begin{tcolorbox}
\textbf{Principle 4: Adaptive Personalization} \\
The agent should learn and predict the user's evolving preferences and abilities, and respect them while executing tasks.
\end{tcolorbox}

The fourth principle is centered around personalization, demanding an agent to be knowledgeable about the user it is interacting with, and to use its knowledge to make decisions while executing tasks for them.
Personalization can be operationally defined as the ability for an AI agent to automatically disambiguate a user's intent, without having to directly ask for an active intervention from them or having a conversation with them.
This can drastically reduce friction in user experiences powered by an agent, since the agent will not need to delay execution of a task for disambiguation reasons, unless strictly needed.

\begin{tcolorbox}
\textbf{Principle 5: Operational Transparency} \\
The agent should expose its inputs, reasoning, plans and past actions to the user at a depth suitable to inform oversight and to build trust.
\end{tcolorbox}

The fifth principle requires an agent to be able to expose its internals to a user, enough for the user to understand how the agent is operating or going to operate in the task of interest.
Transparency is a well-known and studied capability, fundamental for human-AI interaction~\citep{doshi2017towards}.
Despite being critical for other types of AI systems as well, systems based on agents autonomously execute actions on behalf of a user, which might amplify the consequences of their mistakes.
Thus, AI agents potentially need higher transparency to be appropriately monitored or to gain a user's trust.

\begin{tcolorbox}
\textbf{Principle 6: Proactive Safety} \\
The agent should pre-emptively prevent harm to people, data, or property, enforcing privacy, security and ethical constraints before and during execution.
\end{tcolorbox}

The sixth principle frames safety, an essential characteristic that agents should possess, as a capability.
When building a potentially dangerous system such as an agent, its designer should prioritize safety as a first-class requirement for a user to be able to operate the system.
Safety plays a crucial role in establishing trust in the experience powered by the agent, an essential aspect of its usability.
Framing safety as a capability is connected to recent frameworks and approaches proposed in the AI safety community, such as \emph{scalable oversight}~\citep{amodei2016concrete,bowman2022measuring} and \emph{AI control}~\citep{greenblatt2024ai}: an agent designer should consider the idea of creating models with the capability to ensure both their own but also another model's safety in a scalable manner.

\begin{figure}[t]
    \centering
  \includegraphics[width=\linewidth]{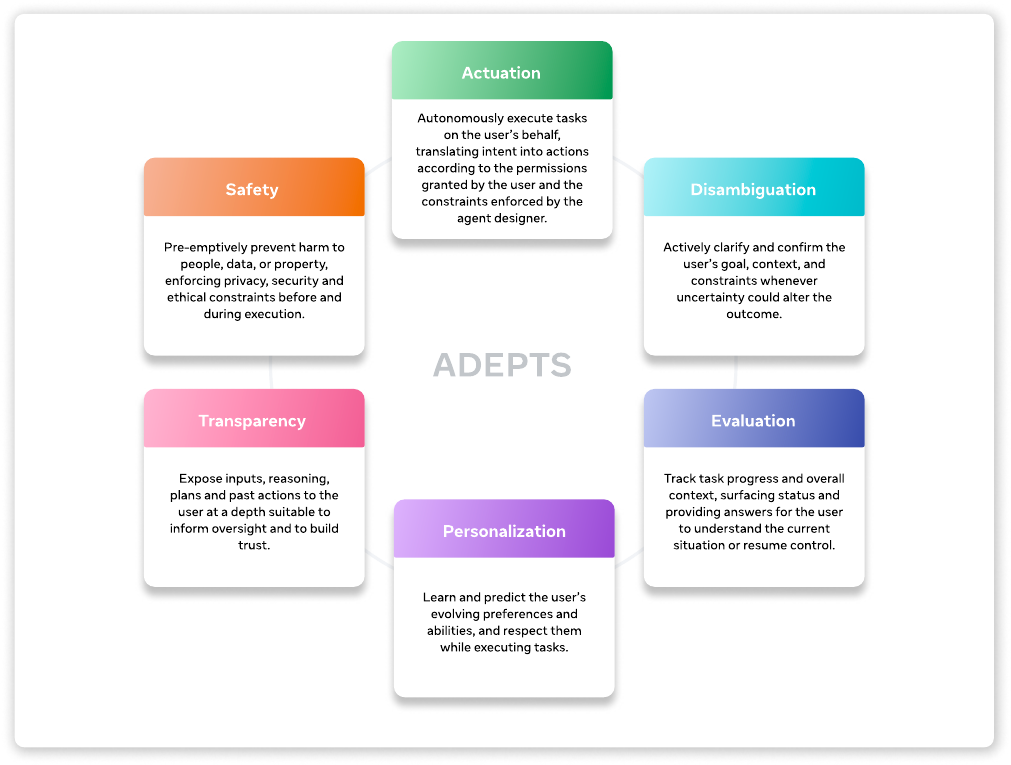}
    \caption{A summary of the ADEPTS capability framework. ADEPTS features six user-facing capabilities, derived from principles for human-centered agent design. 
    }
    \label{fig:adepts}
\end{figure}

\section{The ADEPTS Capability Framework}
To each one of the principles described above corresponds an agent capability.
Taken together, these capabilities constitute a framework that we name ADEPTS (Actuation, Disambiguation, Evaluation, Personalization, Transparency, and Safety).
The intent of ADEPTS is to act as a guideline to suggest functional requirements for creating user experiences based on AI agents.
Note that the ADEPTS framework does not focus on non-functional requirements, such as agent speed and responsiveness.
However, as with software development, non-functional requirements typically play an important role in crafting usable experiences~\citep{chung2012non} and should be considered by an agent designer alongside the ADEPTS capabilities during agent development.

\textbf{Product design considerations.}~~
For product designers, ADEPTS can offer a capability \emph{compass}, as opposed to a set of interface templates.
It tells a product team that an agent should, for example, make its next moves transparent, but it deliberately does not prescribe on \emph{how}, whether that should become a voice narration, a text in a side-bar, or another type of visualization.
Nonetheless, each guiding principle provided by ADEPTS can act as a prompt for a family of design interventions, useful in the process of developing a new experience.
By separating the promise from the presentation, ADEPTS supports creativity while ensuring that different aspects of the design iteration process still trace back to six user-facing obligations from the agent side.

\textbf{Agent design considerations.}~
The ADEPTS framework explicitly gives guidelines for the externally observable behaviors an agent should support, without dictating any particular model architecture, prompt strategy, tool list, or planning procedure that should be used.
It prescribes \emph{what} an agent should be able to do, as measured by an external observer (i.e., a user), not \emph{how} an agent should be able to do it.
For instance, an agent designer may implement disambiguation capabilities using fine-tuned intent classifiers, off-the-shelf LLMs, or any composite system of models. 
ADEPTS is agnostic to this, as long as ambiguous goals trigger an appropriate clarification event. 
To implement different capabilities, an agent can be as modular as deemed appropriate by an agent designer, with each module potentially covering different aspects of a single capability.

\textbf{Capabilities interaction.}~~
While it can often be conceptually and pragmatically useful to think of each capability independently, a crucial responsibility for both product and agent designers is to integrate different capabilities together into a unified system.
Moreover, from an agent design perspective, there might be inherent tradeoffs between the different capabilities. 
For instance, a better performing agent architecture might be more black box, with actuation competency being at stake with transparency; or, better safety guarantees might imply a degradation of absolute agent actuation capability.

\section{Capability Tiers}
We now provide a set of guidelines, in the form of capability tiers related to each of the ADEPTS capabilities.
For each capability, we outline one or more criteria that, taken together, provide a characterization of the competency of an agent regarding that capability.  
We provide a visual summary in Figure~\ref{fig:adepts}.
With the goal of using a language agreed upon among AI researchers, we refer to standard categorizations previously proposed in the community whenever it is possible.
Note that capability tiers are \emph{progressive}: to say that an agent possesses ADEPTS-Tier-4 capabilities in terms of prompt complexity for actuation, the agent should be able to score appropriately high on benchmarks measuring prompt complexity for actuation as prescribed by ADEPTS for all tiers up to the fourth.

\textbf{Illustrative agent types.}~~
For each capability tier, we provide concrete examples of what they might imply in terms of supported agent behavior for illustrative types of agents.
We focus on four types of agent that are gaining increasing popularity both in AI research and product development:
\begin{itemize}
    \item \emph{Computer Use Agents}: they use web browsers, desktop and mobile operating systems~\citep{gao2024generalist,jiang2025comprehensive,qin2025uitars}. They have gained popularity in products for use cases such as workflow automation and online shopping~\citep{openai_operator,anthropic_3_5_models,project_mariner}.
    \item \emph{Coding Agents}: they use code generation and execution abilities to generate code bases as artifacts, developing and testing software~\citep{roziere2023code,liu2024large}. They powered a first wave of products relying on LLM-based AI agents~\citep{cursor_agent,github_copilot}.
    \item \emph{Search Agents}: they rely on search engines and other tools for gathering, aggregating and processing information~\citep{huang2025deep}. A number of products, mostly relying on chat interfaces, have popularized their use, ranging from lightweight web research to gather timely information~\citep{gemini2025technical} to deep research which generate an in-depth report~\citep{openai_deep_research,perplexity_deep_research}.
    \item \emph{Humanoid Agents}: they combine humanoid robots with modern AI agents based on LLMs and their variations~\citep{bjorck2025gr00t,kim2024openvla}. Despite not having yet the broad commercialization and maturity of the previous categories of agents, they are the main focus of a number of well-known product efforts~\citep{figure_helix,tesla_optimus_gen2}.
\end{itemize}

\subsection{Actuation}
\begin{figure}[t!]
    \centering
    \includegraphics[width=\linewidth]{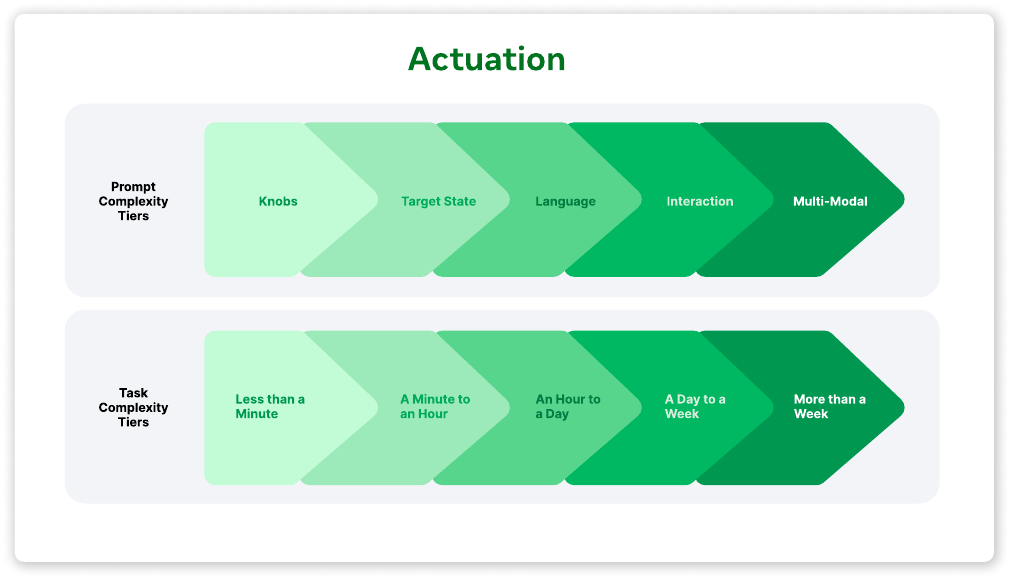}
    \caption{Visual summary of actuation tiers, across prompt complexity and task complexity. Stronger agents should be able to handle higher tiers in the hierarchy.}
    \label{fig:actuation-diagram}
\end{figure}

\subsubsection{Prompt complexity tiers}
Any effective actuation attempt starts with understanding what the user wants the actuation process to actually do.
The more capable an agent is, the more ways it should support to be prompted to achieve a certain actuation objective.
We thus consider tiers of prompt complexity, which give a user increasing flexibility in their way of specifying a goal to the agent.

\textit{Summary of Prompt Complexity tiers for Actuation}: Knobs (Tier 1), Target state (Tier 2), Language (Tier 3), Interactions (Tier 4), Omni-modal (Tier 5). 

\textbf{Tier 1: Knobs. }~~
The agent can receive task instructions in the form of textual or numerical parameters, featuring a finite set of configuration options, which behave similarly to knobs in a physical interface.

\begin{tcolorbox}
\scriptsize
\textbf{Actuation - Prompt Complexity Tier 1 (Knobs)} \\
\rule{\linewidth}{0.8pt} 
\\~\\
\textbf{Computer Use Agent Example} \\
The user can ask the agent to book a flight by choosing from a limited number of dates. \\
\rule{\linewidth}{0.1pt}
\textbf{Coding Agent Example} \\
The user can ask the agent to write a chosen number of unit tests for an existing codebase. \\
\rule{\linewidth}{0.1pt} 
\textbf{Search Agent Example} \\
The user can ask the agent to track the evolution of a set of news topics over time. \\
\rule{\linewidth}{0.1pt} 
\textbf{Humanoid Agent Example} \\
The user can adjust the robot's walking speed by selecting from a set of speed levels.
\end{tcolorbox}

\textbf{Tier 2: Target state. }~~
The agent can receive task instructions in the form of a target state to reach, or of observations that can be linked to an intended target state.  
\begin{tcolorbox}
\scriptsize
\textbf{Actuation - Prompt Complexity Tier 2 (Target state)} \\
\rule{\linewidth}{0.8pt} 
\\~\\
\textbf{Computer Use Agent Example} \\
The user can provide a screenshot of a desired desktop layout, and the agent arranges the icons accordingly. \\
\rule{\linewidth}{0.1pt}
\textbf{Coding Agent Example} \\
The user can provide a wireframe of a web page, and the agent generates the corresponding HTML and CSS. \\
\rule{\linewidth}{0.1pt} 
\textbf{Search Agent Example} \\
The user can provide a draft of a research paper, and the agent finds and adds relevant citations. \\
\rule{\linewidth}{0.1pt} 
\textbf{Humanoid Agent Example} \\
The user can provide a photo of a room setup, and the agent rearranges the furniture to match the configuration.
\end{tcolorbox}

\textbf{Tier 3: Language. }~~
The agent can receive task instruction in the form of open-ended language.
\nopagebreak
\begin{tcolorbox}
\scriptsize
\textbf{Actuation - Prompt Complexity Tier 3 (Language)} \\
\rule{\linewidth}{0.8pt} 
\\~\\
\textbf{Computer Use Agent Example} \\
The user can ask the agent to book any flight from an airline app as described by precise instructions in natural language.
\rule{\linewidth}{0.1pt}
\textbf{Coding Agent Example} \\
The user can ask the agent to refactor a piece of code to improve its efficiency using natural language instructions. \\
\rule{\linewidth}{0.1pt} 
\textbf{Search Agent Example} \\
The user can ask the agent to find articles related to climate change impacts on agriculture using a natural language query.
 \\
\rule{\linewidth}{0.1pt} 
\textbf{Humanoid Agent Example} \\
The user can instruct the robot to clean the living room by describing the tasks in natural language.
\end{tcolorbox}

\textbf{Tier 4: Interactions.}~~
The agent can receive task instructions in the form of interactions. For instance, it could be a set of previous attempts at achieving a goal, or an incomplete attempt combined with a new goal to achieve (i.e., online task reassignment or correction). \nopagebreak

\begin{tcolorbox}
\scriptsize
\textbf{Actuation - Prompt Complexity Tier 4 (Interactions)} \\
\rule{\linewidth}{0.8pt} 
\\~\\
\textbf{Computer Use Agent Example} \\
The agent can be guided to set up a new email account by showing it previous account setups and indicating the name for the new account. \\
\rule{\linewidth}{0.1pt}
\textbf{Coding Agent Example} \\
The agent can be guided to fix a bug by giving it access to past debugging sessions and applying similar solutions to a new issue. \\
\rule{\linewidth}{0.1pt} 
\textbf{Search Agent Example} \\
The agent can refine its search strategy by interacting with the user, who provides feedback on the accuracy of the results. \\
\rule{\linewidth}{0.1pt} 
\textbf{Humanoid Agent Example} \\
The agent can learn to perform a new household task by observing a human perform it and then attempting it with real-time feedback.
\end{tcolorbox}

\textbf{Tier 5: Omni-modal.}~~
The agent can receive any combination of prompting modalities as an input, in any order. This includes incomplete examples, natural language, visual cues, and other modalities. 

\begin{tcolorbox}
\scriptsize
\textbf{Actuation - Prompt Complexity Tier 5 (Omni-modal)} \\
\rule{\linewidth}{0.8pt} 
\\~\\
\textbf{Computer Use Agent Example} \\
The agent can use a combination of voice commands, touch gestures, and eye-tracking to open and operate applications. \\
\rule{\linewidth}{0.1pt}
\textbf{Coding Agent Example} \\
The agent can receive a combination of a code snippet, a voice command, and a flowchart to implement an algorithm. \\
\rule{\linewidth}{0.1pt} 
\textbf{Search Agent Example} \\
The user can use a combination of text input, voice, and visual cues to conduct a search on a multifaceted topic. \\
\rule{\linewidth}{0.1pt} 
\textbf{Humanoid Agent Example} \\
The agent can arrange a room by combining voice instructions, a seating chart, and verbal feedback from the user.
\end{tcolorbox}

\subsubsection{Task complexity tiers}
As a way to measure the capability of the agent to solve more and more complex tasks, regardless of how it is instructed to do so, we consider task execution time.
In particular, we follow recent work tracking progress in agent capabilities, and consider different tiers depending on \emph{how much time it would take a competent human} to solve the tasks that the agent is able to solve~\citep{kwa2025measuring}.
Using human execution time allows to compare different types of agents on the same capability scale.
Note that we only consider the \emph{active execution time}, which ignores the time the human would spend passively waiting during task execution.

\textit{Summary of Task Complexity tiers for Actuation}: Less than a minute (Tier 1), From a minute to a hour (Tier 2), From a hour to a day (Tier 3), From a day to a week (Tier 4), More than a week (Tier 5).

\textbf{Tier 1: Less than a minute.} ~~
The agent can complete tasks that would take a less than a minute. \nopagebreak
\begin{tcolorbox}
\scriptsize
\textbf{Actuation - Task Complexity Tier 1 (Less than a minute)} \\
\rule{\linewidth}{0.8pt} 
\\~\\
\textbf{Computer Use Agent Example} \\
The agent can quickly scroll through a webpage to locate and highlight a specific section of text. \\
\rule{\linewidth}{0.1pt}
\textbf{Coding Agent Example} \\
The agent can format a block of code according to a specific style guide. \\
\rule{\linewidth}{0.1pt} 
\textbf{Search Agent Example} \\
The agent can retrieve the current weather information for a specified location. \\
\rule{\linewidth}{0.1pt} 
\textbf{Humanoid Agent Example} \\
The agent can pick up a dropped item and place it on a table.
\end{tcolorbox}

\textbf{Tier 2: From a minute to a hour.} ~~
The agent can complete tasks that would take from a minute to a hour.
\begin{tcolorbox}
\scriptsize
\textbf{Actuation - Task Complexity Tier 2 (From a minute to a hour)} \\
\rule{\linewidth}{0.8pt} 
\\~\\
\textbf{Computer Use Agent Example} \\
The agent can fill out an online form with user-provided information and submit it. \\
\rule{\linewidth}{0.1pt}
\textbf{Coding Agent Example} \\
The agent can write a simple script to automate a repetitive task, such as renaming files in a directory. \\
\rule{\linewidth}{0.1pt} 
\textbf{Search Agent Example} \\
The agent can compile a list of recent news articles on a specific topic. \\
\rule{\linewidth}{0.1pt} 
\textbf{Humanoid Agent Example} \\
The agent can rearreange a number of objects in small room, navigating around obstacles.
\end{tcolorbox}

\textbf{Tier 3: From a hour to a day.} ~~
The agent can complete tasks that would take from a hour to a day. \nopagebreak
\begin{tcolorbox}
\scriptsize
\textbf{Actuation - Task Complexity Tier 3 (From a hour to a day)} \\
\rule{\linewidth}{0.8pt} 
\\~\\
\textbf{Computer Use Agent Example} \\
The agent can compare prices and features of products across multiple e-commerce sites and compile a list of the best options. \\
\rule{\linewidth}{0.1pt}
\textbf{Coding Agent Example} \\
The agent can develop a basic web application with user authentication features. \\
\rule{\linewidth}{0.1pt} 
\textbf{Search Agent Example} \\
The agent can conduct a detailed market analysis report, including data visualization. \\
\rule{\linewidth}{0.1pt} 
\textbf{Humanoid Agent Example} \\
The agent can assemble a piece of furniture using provided instructions and tools.
\end{tcolorbox}

\textbf{Tier 4: From a day to a week.} ~~
The agent can complete tasks that would take from a day to a week. \nopagebreak
\begin{tcolorbox}
\scriptsize
\textbf{Actuation - Task Complexity Tier 4 (From a hour to a day)} \\
\rule{\linewidth}{0.8pt} 
\\~\\
\textbf{Computer Use Agent Example} \\
 The agent can edit a complex video project using professional editing software, incorporating effects, transitions, and audio synchronization to produce a polished final cut. \\
\rule{\linewidth}{0.1pt}
\textbf{Coding Agent Example} \\
The agent can refactor a large codebase to improve performance and maintainability. \\
\rule{\linewidth}{0.1pt} 
\textbf{Search Agent Example} \\
The agent can perform a comprehensive literature review on a scientific topic. \\
\rule{\linewidth}{0.1pt} 
\textbf{Humanoid Agent Example} \\
The agent can assist in setting up an entire event, including arranging furniture, decorations, and equipment.
\end{tcolorbox}

\textbf{Tier 5: More than a week.} ~~
The agent can complete tasks that would take more than a week. \nopagebreak
\begin{tcolorbox}
\scriptsize
\textbf{Actuation - Task Complexity Tier 5 (More than a week)} \\
\rule{\linewidth}{0.8pt} 
\\~\\
\textbf{Computer Use Agent Example} \\
 The agent can create a detailed digital animation using advanced software, handling tasks such as modeling, rigging, texturing, and rendering to produce a high-quality animated short film. \\
\rule{\linewidth}{0.1pt}
\textbf{Coding Agent Example} \\
The agent can develop a full-featured software application, including design, implementation, and testing. \\
\rule{\linewidth}{0.1pt} 
\textbf{Search Agent Example} \\
The agent can conduct an in-depth research project, including data collection, analysis, and redaction of a rigorous write-up. \\
\rule{\linewidth}{0.1pt} 
\textbf{Humanoid Agent Example} \\
The agent can autonomously participate in a construction project, assisting with tasks such as material handling and assembly over several weeks.
\end{tcolorbox}

\subsection{Disambiguation} 
\begin{figure}[t!]
    \centering
    \includegraphics[width=\linewidth]{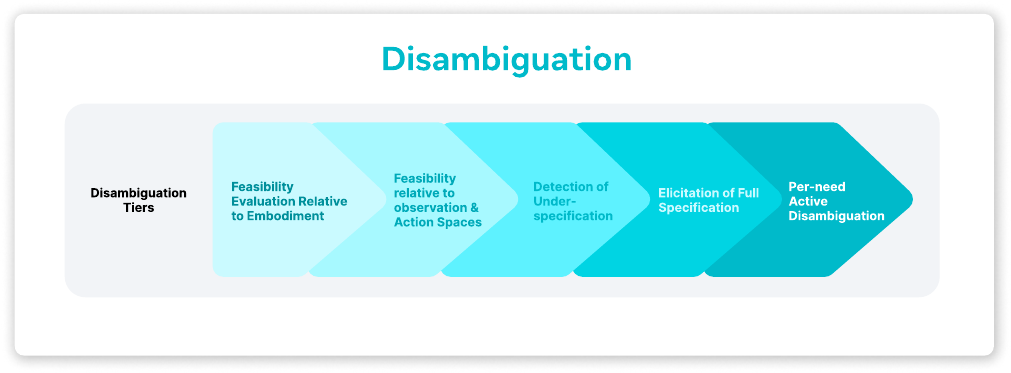}
    \caption{Visual summary of disambiguation tiers. Stronger agents should be able to handle higher tiers in the hierarchy.}
    \label{fig:disambiguation-diagram}
\end{figure}
\subsubsection{Disambiguation tiers}
We consider different tiers of disambiguation capabilities, featuring increasing levels of ambiguity and of active involvement of the agent in the disambiguation process.

\textit{Summary of Disambiguation tiers}: Feasibility evaluation with respect to embodiment (Tier 1), Feasibility evaluation with respect to observation and action space (Tier 2), Underspecification detection (Tier 3), Elicitation of full specification (Tier 4), Per-need active disambiguation (Tier 5).

\textbf{Tier 1: Feasibility evaluation with respect to embodiment.}~~
The agent is able to evaluate whether it is feasible to execute a given task with its embodiment type and to refuse to execute tasks that are impossible. \nopagebreak

\begin{tcolorbox}
\scriptsize
\textbf{Disambiguation - Disambiguation Tier 1 (Feasibility evaluation with respect to embodiment)} \\
\rule{\linewidth}{0.8pt} 
\\~\\
\textbf{Computer Use Agent Example} \\
The agent will refuse if the user directly asks the agent to fold their laundry.  \\
\rule{\linewidth}{0.1pt}
\textbf{Coding Agent Example} \\
The agent will refuse if the user asks it to physically assemble a computer. \\
\rule{\linewidth}{0.1pt} 
\textbf{Search Agent Example} \\
The agent will refuse if the user asks it to physically retrieve a book from a library. \\
\rule{\linewidth}{0.1pt} 
\textbf{Humanoid Agent Example} \\
The agent will refuse if the user asks it to refactor their codebase.
\end{tcolorbox}

\textbf{Tier 2: Feasibility evaluation with respect to observation and action space.}~~
The agent is able to evaluate whether it is feasible to execute a given task using its observation space and its action space and to refuse to execute tasks that are impossible. \nopagebreak

\begin{tcolorbox}
\scriptsize
\textbf{Disambiguation - Disambiguation Tier 2 (Feasibility evaluation w.r.t. observation and action space)} \\
\rule{\linewidth}{0.8pt} 
\\~\\
\textbf{Computer Use Agent Example} \\
 The agent will refuse if the user asks to zoom in if the agent has no support for zooming actions. \\
\rule{\linewidth}{0.1pt}
\textbf{Coding Agent Example} \\
The agent will refuse if the user asks it to compile code in a language it does not support.  \\
\rule{\linewidth}{0.1pt} 
\textbf{Search Agent Example} \\
The agent will refuse if the user asks it to access a database that is outside its network permissions. \\
\rule{\linewidth}{0.1pt} 
\textbf{Humanoid Agent Example} \\
The agent will refuse if the user asks it to fly without using any tool.
\end{tcolorbox}

\textbf{Tier 3: Underspecification detection.}~~
The agent is able to detect that an instruction misses required information.\nopagebreak
\begin{tcolorbox}
\scriptsize
\textbf{Disambiguation - Disambiguation Tier 3 (Underspecification detection)} \\
\rule{\linewidth}{0.8pt} 
\\~\\
\textbf{Computer Use Agent Example} \\
 The agent recognizes "book a flight" as an instruction requiring more information to be a valid goal. \\
\rule{\linewidth}{0.1pt}
\textbf{Coding Agent Example} \\
The agent detects that "optimize the code" is underspecified and requires details on which aspects to optimize. \\
\rule{\linewidth}{0.1pt} 
\textbf{Search Agent Example} \\
The agent identifies "find articles" as underspecified, missing a particular topic or keywords.

\rule{\linewidth}{0.1pt} 
\textbf{Humanoid Agent Example} \\
The agent identifies "organize the bookshelf" as underspecified, noting that it needs more information on the desired organization method, such as by genre or author.
\end{tcolorbox}

\textbf{Tier 4: Elicitation of full specification.} ~~
The agent is able to conduct a full conversation with a user until their goal is fully specified.  \nopagebreak
\begin{tcolorbox}
\scriptsize
\textbf{Disambiguation - Disambiguation Tier 4 (Elicitation of full specification)} \\
\rule{\linewidth}{0.8pt} 
\\~\\
\textbf{Computer Use Agent Example} \\
 After receiving the instruction "book a flight," the agent chats with the user to understand from where to where, at what time, with which airline, and what ticket class. \\
\rule{\linewidth}{0.1pt}
\textbf{Coding Agent Example} \\
The agent engages in a dialogue to clarify "create a user interface," asking about design preferences, features, and target platforms. \\
\rule{\linewidth}{0.1pt} 
\textbf{Search Agent Example} \\
The agent converses with the user to refine "research climate change," asking for specific aspects or regions of interest. \\
\rule{\linewidth}{0.1pt} 
\textbf{Humanoid Agent Example} \\
The agent interacts with the user to fully specify "prepare a meal," asking about dietary preferences, ingredients, and serving time.
\end{tcolorbox}

\textbf{Tier 5: Per-need active disambiguation.} ~~
While actuating, the agent is able to detect there is need for disambiguation and to elicit required information from the user. This can both mean that the agent starts actuating from instructions that are not fully disambiguated, and initiates a disambiguation interaction when needed, or that it seeks disambiguation when an unexpected system state renders some part of the user-provided task specification invalid.   

\begin{tcolorbox}
\scriptsize
\textbf{Disambiguation - Disambiguation Tier 5 (Per-need active disambiguation)} \\
\rule{\linewidth}{0.8pt} 
\\~\\
\textbf{Computer Use Agent Example} \\
 The agent asks the user what other airline they prefer when it realizes that the website of the originally prescribed airline for booking a flight ticket is down. \\
\rule{\linewidth}{0.1pt}
\textbf{Coding Agent Example} \\
The agent pauses and asks the user for alternative libraries when it encounters a deprecated library during code execution. \\
\rule{\linewidth}{0.1pt} 
\textbf{Search Agent Example} \\
The agent requests additional keywords or sources when it finds that the initial search results are too broad or irrelevant. \\
\rule{\linewidth}{0.1pt} 
\textbf{Humanoid Agent Example} \\
The agent seeks clarification on cleaning priorities when it encounters unexpected obstacles or changes in the room layout.
\end{tcolorbox}

\subsection{Evaluation} 
\subsubsection{Evaluation mode tiers}
We consider different tiers for what type of evaluation an agent can provide.
They range from descriptive evaluation capabilities to interactive and predictive approaches to evaluation.
Advanced evaluation capabilities are not only extremely useful for powering trustworthy user experiences, but can also be instrumental during agent training for obtaining automated evaluation signals.

\begin{figure}[t!]
    \centering
    \includegraphics[width=\linewidth]{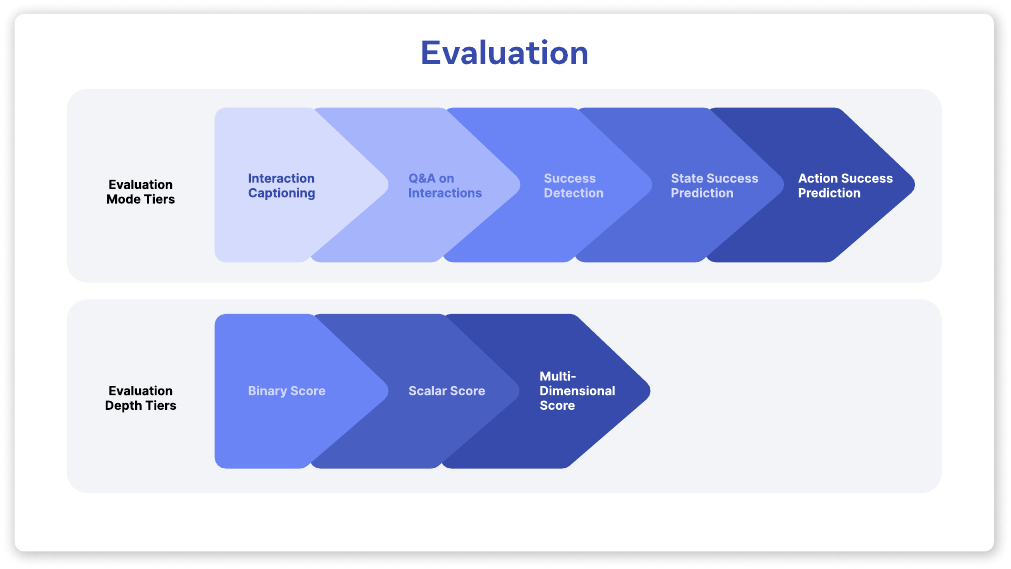}
    \caption{Visual summary of evaluation tiers, in terms of evaluation mode and evaluation depth. Stronger agents should be able to handle higher tiers in the hierarchy.}
    \label{fig:evaluation-diagram}
\end{figure}

\textit{Summary of Evaluation Mode tiers for Evaluation}: Interaction captioning (Tier 1), Question-answering on interactions (Tier 2), Success detection (Tier 3), State-based success prediction (Tier 4), Action-based success prediction (Tier 5). 

\textbf{Tier 1: Interaction captioning.}~~
The agent is able to caption and narrate interactions with the system at hand. \nopagebreak
\begin{tcolorbox}
\scriptsize
\textbf{Evaluation - Evaluation Mode Tier 1 (Interaction captioning)} \\
\rule{\linewidth}{0.8pt} 
\\~\\
\textbf{Computer Use Agent Example} \\
The agent provides a detailed description of the steps taken during an online shopping session, such as items viewed, added to the cart, and any filters applied. \\
\rule{\linewidth}{0.1pt}
\textbf{Coding Agent Example} \\
The agent generates a summary of recent changes made to a codebase, including commits, merges, and any conflicts resolved. \\
\rule{\linewidth}{0.1pt} 
\textbf{Search Agent Example} \\
The agent describes the sequence of search queries and filters applied during a research session (e.g., by exposing its thinking trace). \\
\rule{\linewidth}{0.1pt} 
\textbf{Humanoid Agent Example} \\
The agent provides a description of the state of a room, describing how the various objects in the room are organized.
\end{tcolorbox}

\textbf{Tier 2: Question-answering on interactions.}~~
The agent is able to answer a user's questions on the relevant state of the world or on previous interactions. 

\begin{tcolorbox}
\scriptsize
\textbf{Evaluation - Evaluation Mode Tier 2 (Question-answering on interactions)} \\
\rule{\linewidth}{0.8pt} 
\\~\\
\textbf{Computer Use Agent Example} \\
The agent answers questions about the current state of a document editing program, such as "What formatting options are available for this text?" \\
\rule{\linewidth}{0.1pt}
\textbf{Coding Agent Example} \\
The agent responds to inquiries about how the codebase works. \\
\rule{\linewidth}{0.1pt} 
\textbf{Search Agent Example} \\
The agent can answer questions about the meaning and use of some technical words it used in the answer provided to a search. \\
\rule{\linewidth}{0.1pt} 
\textbf{Humanoid Agent Example} \\
The agent can answer questions about the different objects in a room and how they can be used.
\end{tcolorbox}

\textbf{Tier 3: Success detection.}~~
The agent is able to detect whether a past interaction is efficiently and effectively exhibiting the behavior specified in a user's instructions.
Using the language of reinforcement learning, this is equivalent to the ability of approximating a reward function~\citep{sutton1998reinforcement}. \nopagebreak
\begin{tcolorbox}
\scriptsize
\textbf{Evaluation - Evaluation Mode Tier 3 (Success detection)} \\
\rule{\linewidth}{0.8pt} 
\\~\\
\textbf{Computer Use Agent Example} \\
The agent evaluates whether an online purchase was successfully completed, confirming that the order was placed and payment processed. \\
\rule{\linewidth}{0.1pt}
\textbf{Coding Agent Example} \\
The agent checks if a newly implemented feature passes all unit tests and integrates seamlessly with existing code. \\
\rule{\linewidth}{0.1pt} 
\textbf{Search Agent Example} \\
The agent determines if a search session successfully yielded the desired information. \\
\rule{\linewidth}{0.1pt} 
\textbf{Humanoid Agent Example} \\
The agent assesses whether a completed task, such as setting up a dining table, meets the specified requirements.
\end{tcolorbox}

\textbf{Tier 4: State-based success prediction.}~~
The agent is able to evaluate an incomplete interaction (or the current state of the system) to predict whether it will ultimately lead to achieving the goal or exhibiting the behavior specified in a user's instruction. 
Using the language of reinforcement learning, this is equivalent to the ability of approximating a value function~\citep{sutton1998reinforcement}. \nopagebreak
\begin{tcolorbox}
\scriptsize
\textbf{Evaluation - Evaluation Mode Tier 4: (State-based success prediction.)} \\
\rule{\linewidth}{0.8pt} 
\\~\\
\textbf{Computer Use Agent Example} \\
The agent predicts whether the current configuration of a photo editing project will lead to the desired visual outcome. \\
\rule{\linewidth}{0.1pt}
\textbf{Coding Agent Example} \\
The agent evaluates an ongoing debugging process to predict if the identified solution will resolve the issue. \\
\rule{\linewidth}{0.1pt} 
\textbf{Search Agent Example} \\
The agent predicts whether the information that was found in previous searches will lead to finding comprehensive results for a research topic. \\
\rule{\linewidth}{0.1pt} 
\textbf{Humanoid Agent Example} \\
The agent assesses the current progress of a home renovation project to predict if it will be completed within the planned timeline.
\end{tcolorbox}

\textbf{Tier 5: Action-based success prediction.}~~
The agent is able to evaluate whether a proposed action in a given situation (i.e., system state) is promising to achieve the goal or exhibit the behavior specified in a user's instruction.
Using the language of reinforcement learning, this is equivalent to the ability of approximating an action-value function~\citep{sutton1998reinforcement}.
\begin{tcolorbox}
\textbf{Evaluation - Evaluation Mode Tier 5: (Action-based success prediction)} \\
\scriptsize
\rule{\linewidth}{0.8pt} 
\\~\\
\textbf{Computer Use Agent Example} \\
The agent evaluates whether clicking a specific button in a software application will successfully execute a desired function, such as saving a document. \\
\rule{\linewidth}{0.1pt}
\textbf{Coding Agent Example} \\
The agent predicts whether changing a single line of code will fix a bug without introducing new issues, based on the current code context. \\
\rule{\linewidth}{0.1pt} 
\textbf{Search Agent Example} \\
The agent assesses whether using a different search engine will yield more relevant results for a specific query. \\
\rule{\linewidth}{0.1pt} 
\textbf{Humanoid Agent Example} \\
The agent predicts whether adjusting its grip on an object will improve its ability to lift the object without dropping it.
\end{tcolorbox}

\subsubsection{Evaluation depth tiers}
We consider different evaluation depth tiers.
A more competent agent should provide more precise numerical information about the quality of the interactions and their related predictions.
In reinforcement learning terms, this quantifies how dense and rich, as opposed to sparse, an evaluation signal is~\citep{sutton1998reinforcement}.

\textit{Summary of Evaluation Depth tiers for Evaluation}: Binary score (Tier 1), Scalar score state (Tier 2), Multi-dimensional score (Tier 3). 

\textbf{Tier 1: Binary score.} ~~
The agent is able to produce a binary evaluation score. \nopagebreak
\begin{tcolorbox}
\scriptsize
\textbf{Evaluation - Evaluation Depth Tier 1: (Binary score)} \\
\rule{\linewidth}{0.8pt} 
\\~\\
\textbf{Computer Use Agent Example} \\
The agent evaluates whether a file was successfully uploaded to a cloud service. \\
\rule{\linewidth}{0.1pt}
\textbf{Coding Agent Example} \\
The agent checks if some code successfully implements the desired behavior or not. \\
\rule{\linewidth}{0.1pt} 
\textbf{Search Agent Example} \\
The agent determines if a search query returned any sensible results or not. \\
\rule{\linewidth}{0.1pt} 
\textbf{Humanoid Agent Example} \\
The agent assesses whether a door was successfully opened or not.
\end{tcolorbox}

\textbf{Tier 2: Scalar score.} ~~
The agent is able to provide a scalar evaluation score (e.g., incorporating preferences).
\begin{tcolorbox}
\scriptsize
\textbf{Evaluation - Evaluation Depth Tier 2: (Scalar score)} \\
\rule{\linewidth}{0.8pt} 
\\~\\
\textbf{Computer Use Agent Example} \\
The agent evaluates groceries bought for the user, giving a score from 1 to 10 based on how much they match with the user preferences. \\
\rule{\linewidth}{0.1pt}
\textbf{Coding Agent Example} \\
The agent evaluates the quality of a code refactoring effort, providing a scalar score based on improvements in readability. \\
\rule{\linewidth}{0.1pt} 
\textbf{Search Agent Example} \\
The agent scores the relevance of search results on a scale from 1 to 5, based on how well they match the user's query. \\
\rule{\linewidth}{0.1pt} 
\textbf{Humanoid Agent Example} \\
The agent rates the cleanliness of a room on a scale from 1 to 10.
\end{tcolorbox}

\textbf{Tier 3: Multi-dimensional score.} ~~
The agent is able to provide a multi-dimensional evaluation score considering multiple criteria.

\begin{tcolorbox}
\scriptsize
\textbf{Evaluation - Evaluation Depth Tier 3: (Multi-dimensional score)} \\
\rule{\linewidth}{0.8pt} 
\\~\\
\textbf{Computer Use Agent Example} \\
The agent provides a multi-dimensional score for a video editing project, evaluating criteria such as visual quality, audio synchronization, and rendering speed. \\
\rule{\linewidth}{0.1pt}
\textbf{Coding Agent Example} \\
The agent assesses a software application using a multi-dimensional score, considering factors like functionality, user interface design, and performance. \\
\rule{\linewidth}{0.1pt} 
\textbf{Search Agent Example} \\
The agent evaluates a search session with a multi-dimensional score, considering criteria such as relevance, diversity of sources, and timeliness of information. \\
\rule{\linewidth}{0.1pt} 
\textbf{Humanoid Agent Example} \\
The agent evaluates the setup of a conference room with a multi-dimensional score, considering factors such as seating arrangement, audio-visual equipment functionality, and overall accessibility.
\end{tcolorbox}

\subsection{Personalization} 
\begin{figure}[t!]
    \centering
    \includegraphics[width=\linewidth]{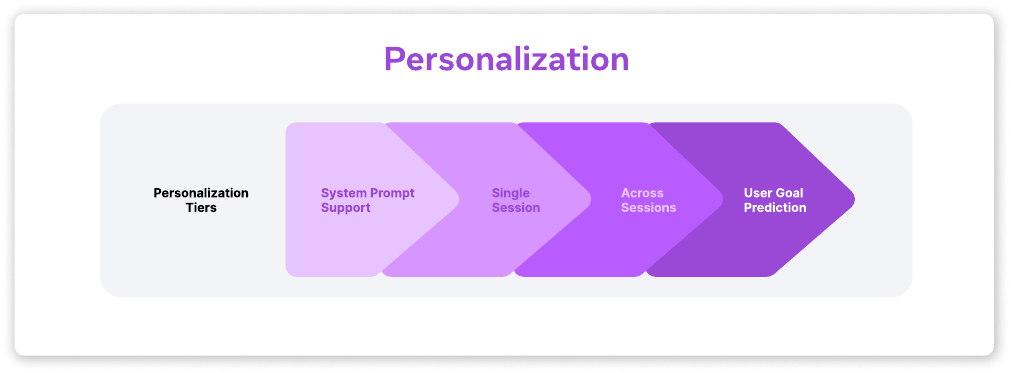}
    \caption{Visual summary of personalization tiers. Stronger agents should be able to handle higher tiers in the hierarchy.}
    \label{fig:personalization-diagram}
\end{figure}
\subsubsection{Personalization tiers}

We consider a number of personalization tiers, going from supporting simple generic instructions (e.g., in text) that can be extracted from sources about the user to agents behaving similarly to recommender system.

\textit{Summary of Personalization tiers}: System prompt support (Tier 1), Single-session (Tier 2), Across-sessions (Tier 3), User goal prediction (Tier 4). 

\textbf{Tier 1: System prompt support.}~~
The agent supports the use of "system prompts", that it can use to receive information about the user (e.g., about their preferences) from other systems or manually from users themselves. \nopagebreak
\begin{tcolorbox}
\scriptsize
\textbf{Personalization - Personalization Tier 1: (System prompt support)} \\
\rule{\linewidth}{0.8pt} 
\\~\\
\textbf{Computer Use Agent Example} \\
The agent uses system prompts to customize the user interface of a software application based on preferences received from a user profile, such as preferred language, theme, and font size. \\
\rule{\linewidth}{0.1pt}
\textbf{Coding Agent Example} \\
The agent applies coding style preferences, such as indentation and naming conventions, based on a configuration file provided by the user. \\
\rule{\linewidth}{0.1pt} 
\textbf{Search Agent Example} \\
The agent tailors search results by incorporating user interests and preferred sources as aggregated from a social media account. \\
\rule{\linewidth}{0.1pt} 
\textbf{Humanoid Agent Example} \\
The agent adjusts its cleaning routine based on user preferences for cleaning products, as specified via manual input.
\end{tcolorbox}
\textbf{Tier 2: Single session.}~~
The agent can infer and use preferences of a user in a single session. \nopagebreak
\begin{tcolorbox}
\scriptsize
\textbf{Personalization - Personalization Tier 2: (Single-session)} \\
\rule{\linewidth}{0.8pt} 
\\~\\
\textbf{Computer Use Agent Example} \\
During a single online shopping session, the agent learns the user's preference for sorting items by price and automatically applies this filter to subsequent product searches. \\
\rule{\linewidth}{0.1pt}
\textbf{Coding Agent Example} \\
While working on a coding project, the agent notices the user's preference for using specific libraries and suggests relevant code snippets and documentation within the same session. \\
\rule{\linewidth}{0.1pt} 
\textbf{Search Agent Example} \\
The agent observes that the user frequently clicks on articles from a particular news source during a session and prioritizes results from that source in real-time. \\
\rule{\linewidth}{0.1pt} 
\textbf{Humanoid Agent Example} \\
The agent learns the user's preferred room temperature and adjusts the thermostat in other rooms following that during a single interaction.
\end{tcolorbox}

\textbf{Tier 3: Across sessions.}~~
The agent can infer and use preferences of a user across sessions, possibly modeling their evolution.
\begin{tcolorbox}
\scriptsize
\textbf{Personalization - Personalization Tier 3: (Across-session)} \\
\rule{\linewidth}{0.8pt} 
\\~\\
\textbf{Computer Use Agent Example} \\
The agent remembers the user's preferred document templates and automatically suggests them in future sessions. \\
\rule{\linewidth}{0.1pt}
\textbf{Coding Agent Example} \\
The agent tracks the user's coding habits and suggests improvements or shortcuts based on patterns observed across multiple sessions. \\
\rule{\linewidth}{0.1pt} 
\textbf{Search Agent Example} \\
The agent builds a profile of the user's research interests over time and tailors search strategies to align with evolving preferences. \\
\rule{\linewidth}{0.1pt} 
\textbf{Humanoid Agent Example} \\
The agent learns the user's daily routine and adjusts its tasks, such as preparing coffee or setting alarms, based on patterns observed over time.
\end{tcolorbox}

\textbf{Tier 4: User goal prediction.}~~
The agent can predict what the next goal or instruction given by the user could be, and act as a recommendation engine for the user. 
\begin{tcolorbox}
\scriptsize
\textbf{Personalization - Personalization Tier 4: (User goal prediction)} \\
\rule{\linewidth}{0.8pt} 
\\~\\
\textbf{Computer Use Agent Example} \\
The agent predicts the user's next task, such as opening a frequently used application. \\
\rule{\linewidth}{0.1pt}
\textbf{Coding Agent Example} \\
The agent anticipates the user's next coding task, such as debugging or testing, and prepares the necessary tools and resources. \\
\rule{\linewidth}{0.1pt} 
\textbf{Search Agent Example} \\
The agent predicts the user's next search query based on recent activity and suggests related topics or articles. \\
\rule{\linewidth}{0.1pt} 
\textbf{Humanoid Agent Example} \\
The agent anticipates the user's need for assistance, such as preparing a meal or organizing a workspace, and offers to help before being asked. \\
\end{tcolorbox}

\subsection{Transparency} 

\begin{figure}[t!]
    \centering
    \includegraphics[width=\linewidth]{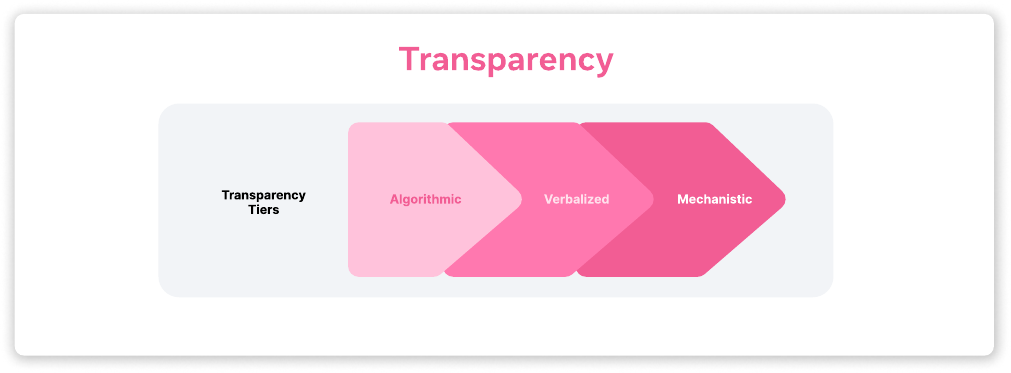}
    \caption{Visual summary of transparency tiers. Stronger agents should be able to handle higher tiers in the hierarchy.}
    \label{fig:transparency-diagram}
\end{figure}

\subsubsection{Transparency tiers}
We define three distinct transparency tiers, with increasing difficulty in how they can be implemented and how deep is the type of information that gets surfaced to the user.

\textit{Summary of Transparency tiers}: Algorithmic (Tier 1), Verbalized (Tier 2), Mechanistic (Tier 3). 

\textbf{Tier 1: Algorithmic.}~~
The agent can passively show how, by construction, information flows into its architecture, also including the system prompts of the models it is using and any other technical aspect (e.g., by showing its own code). 
\begin{tcolorbox}
\scriptsize
\textbf{Transparency - Transparency Tier 1: (Algorithmic)} \\
\rule{\linewidth}{0.8pt} 
\\~\\
\textbf{Computer Use Agent Example} \\
The agent provides access to its source code and system prompts, allowing users to see how it processes data and makes decisions within a software application. \\
\rule{\linewidth}{0.1pt}
\textbf{Coding Agent Example} \\
The agent displays the algorithms and libraries it uses for code analysis, enabling users to understand the underlying logic of its suggestions. \\
\rule{\linewidth}{0.1pt} 
\textbf{Search Agent Example} \\
The agent shows the search algorithms and ranking criteria it employs, giving users insight into how search results are prioritized. \\
\rule{\linewidth}{0.1pt} 
\textbf{Humanoid Agent Example} \\
The agent reveals the control algorithms and sensor data it uses to navigate and interact with its environment, offering transparency into its operational processes. 
\end{tcolorbox}

\textbf{Tier 2: Verbalized.}~~
The agent is able to verbalize, in natural language or code, the rationale behind its actions and its thoughts in a way that can be understood by a user (e.g., via a faithful LLM Chain of Thought~\citep{wei2022chain,lyu2023faithful}). \nopagebreak
\begin{tcolorbox}
\scriptsize
\textbf{Transparency - Transparency Tier 2: (Verbalized)} \\
\rule{\linewidth}{0.8pt} 
\\~\\
\textbf{Computer Use Agent Example} \\
The agent explains in natural language why it prioritized using a certain app instead of another, detailing the criteria and user preferences considered. \\
\rule{\linewidth}{0.1pt}
\textbf{Coding Agent Example} \\
The agent verbalizes its reasoning for suggesting a specific code refactor, outlining the potential benefits and improvements in performance. \\
\rule{\linewidth}{0.1pt} 
\textbf{Search Agent Example} \\
The agent describes its thought process in selecting certain articles as relevant, explaining the keywords and context it considered. \\
\rule{\linewidth}{0.1pt} 
\textbf{Humanoid Agent Example} \\
The agent articulates its decision-making process when choosing a path to navigate a room, explaining how it assessed obstacles and selected the most efficient route.
\end{tcolorbox}

\textbf{Tier 3: Mechanistic.}~~
The agent is able to expose its internal reasoning and processing rationales, as derived from mechanistic inspection of its components (e.g., mechanistic interpretability of its neural networks~\citep{bereska2024mechanistic}). \nopagebreak
\begin{tcolorbox}
\scriptsize
\textbf{Transparency - Transparency Tier 3: (Mechanistic)} \\
\rule{\linewidth}{0.8pt} 
\\~\\
\textbf{Computer Use Agent Example} \\
The agent provides a detailed breakdown of its neural network's decision-making process, showing how different layers and nodes contributed to a specific output. \\
\rule{\linewidth}{0.1pt}
\textbf{Coding Agent Example} \\
The agent exposes the internal workings of its code analysis model, illustrating how it identifies patterns and anomalies in the codebase. \\
\rule{\linewidth}{0.1pt} 
\textbf{Search Agent Example} \\
The agent offers a mechanistic view of its search ranking model, detailing how various features and weights influence the final ranking of search results. \\
\rule{\linewidth}{0.1pt} 
\textbf{Humanoid Agent Example} \\
The agent reveals how its neural network processes visual inputs to recognize objects. It shows how specific neurons activate in response to different shapes and colors, explaining how these activations lead to decisions about object manipulation.
\end{tcolorbox}

\subsection{Safety} 
\begin{figure}[t!]
    \centering
    \includegraphics[width=\linewidth]{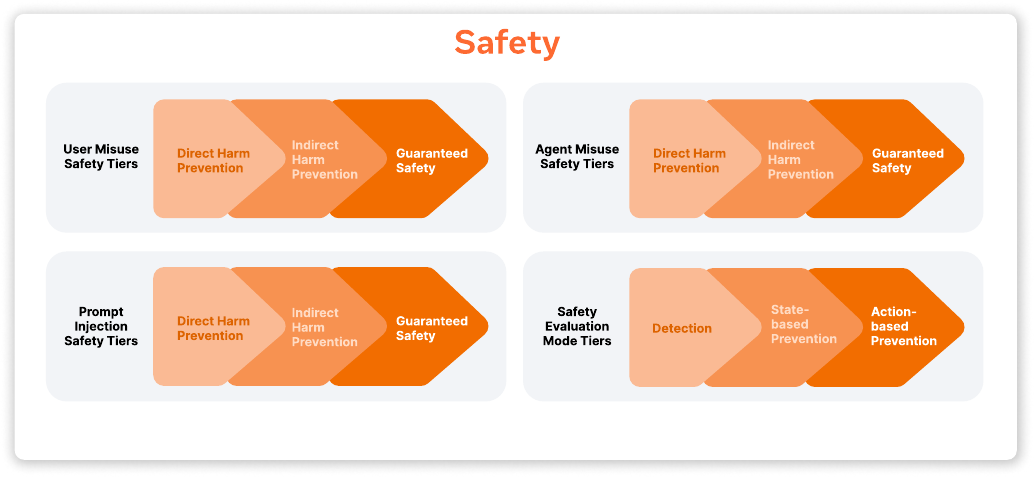}
    \caption{Visual summary of safety tiers, in terms of user misuse, agent misbehavior, prompt injection, and safety evaluation mode. Stronger agents should be able to handle higher tiers in the hierarchy.}
    \label{fig:safety-diagram}
\end{figure}
For creating different tier categories for agent safety, we follow the categorization recently introduced by \citet{openai_operator_system_card} and \citet{kuntz2025harm}.
This categorization follows standard abstractions around input-output safety, which model user safety and agent safety separately~\citep{sharma2025constitutional,inan2023llama}.
In addition, we consider different safety evaluation mode tiers, with increasing levels of proactivity in detecting potential safety violations.

\subsubsection{User misuse safety tiers}
We consider user misuse as one of the axis to evaluate the safety of agent-based systems.
User misuse safety considers the harm that can be caused by a user in an intentional way by instructing the system.
The different tiers  are related to how powerful it is the system at detecting and preventing user misuse.

\textit{Summary of User Misuse Safety tiers}: Direct harm (Tier 1), Indirect harm (Tier 2), Guaranteed safety (Tier~3). 

\textbf{Tier 1: Direct harm.}~~
The agent is able to detect or prevent the execution of tasks that could cause direct harm if completed, such as hacking into a system or causing physical damage. \nopagebreak
\begin{tcolorbox}
\scriptsize
\textbf{Safety - User Misuse Safety Tier 1: (Direct harm)} \\
\rule{\linewidth}{0.8pt} 
\\
\textbf{Computer Use Agent Example} \\
An agent that refuses to execute commands that would delete critical system files. \\
\rule{\linewidth}{0.1pt}
\textbf{Coding Agent Example} \\
An agent that prevents the creation of scripts designed to overload a server. \\
\rule{\linewidth}{0.1pt} 
\textbf{Search Agent Example} \\
An agent that filters out search results related to illegal activities. \\
\rule{\linewidth}{0.1pt} 
\textbf{Humanoid Agent Example} \\
An agent that does not follow instructions to physically harm a person. 
\end{tcolorbox}

\textbf{Tier 2: Indirect harm.}~~
The agent is able to detect or prevent tasks that could indirectly lead to harm, such as gathering information for harmful purposes. 
\begin{tcolorbox}
\scriptsize
\textbf{Safety - User Misuse Safety Tier 2: (Indirect harm)} \\
\rule{\linewidth}{0.8pt} 
\\
\textbf{Computer Use Agent Example} \\
An agent that blocks access to tutorials on creating malware. \\
\rule{\linewidth}{0.1pt}
\textbf{Coding Agent Example} \\
An agent that refuses to compile code snippets that could be used for phishing. \\
\rule{\linewidth}{0.1pt} 
\textbf{Search Agent Example} \\
An agent that does not return results for queries on building weapons. \\
\rule{\linewidth}{0.1pt} 
\textbf{Humanoid Agent Example} \\
An agent that ignores requests to gather sensitive personal information. 
\end{tcolorbox}

\textbf{Tier 3: Guaranteed safety.}~~
The agent is able to ensure that, under some assumptions and with a certain probabilistic confidence~\citep{bengio2025superintelligent}, instructions do not lead to any harm, directly or indirectly. \nopagebreak
\begin{tcolorbox}
\scriptsize
\textbf{Safety - User Misuse Safety Tier 3: (Guaranteed safety)} \\
\rule{\linewidth}{0.8pt} 
\\
\textbf{Computer Use Agent Example} \\
An agent that only allows operations that are deemed as safe with 95\% confidence according to a rubric. \\
\rule{\linewidth}{0.1pt}
\textbf{Coding Agent Example} \\
An agent that only compiles code from libraries it is 99\% they are safe. \\
\rule{\linewidth}{0.1pt} 
\textbf{Search Agent Example} \\
An agent that provides results only from websites safe with 99\% confidence. \\
\rule{\linewidth}{0.1pt} 
\textbf{Humanoid Agent Example} \\
An agent that only manipulates objects when it is 99\% confident that it can manipulate them without breaking them. 
\end{tcolorbox}

\subsubsection{Agent misbehavior safety tiers}
Agent misbehavior safety concerns detecting and preventing harm coming from initiative or incompetency of the agent itself.
We follow similar tiers as the ones established for user misuse.

\textit{Summary of User Misuse Safety tiers}: Direct harm (Tier 1), Indirect harm (Tier 2), Guaranteed safety (Tier~3). 

\textbf{Tier 1: Prevention of direct harm.}~~
The agent is able to prevent actions that could directly cause harm due to misbehavior. \nopagebreak
\begin{tcolorbox}
\scriptsize
\textbf{Safety - Agent Misbehavior Safety Tier 1: (Direct harm)} \\
\rule{\linewidth}{0.8pt} 
\\
\textbf{Computer Use Agent Example} \\
An agent that does not execute actions that could crash a system. \\
\rule{\linewidth}{0.1pt}
\textbf{Coding Agent Example} \\
An agent that avoids generating code that could lead to data corruption. \\
\rule{\linewidth}{0.1pt} 
\textbf{Search Agent Example} \\
An agent that does not provide links to harmful websites. \\
\rule{\linewidth}{0.1pt} 
\textbf{Humanoid Agent Example} \\
An agent that does not perform actions that could physically harm users. 
\end{tcolorbox}

\textbf{Tier 2: Prevention of indirect harm.}~~
The agent is able to prevent actions that could indirectly lead to harm through misbehavior. 
\begin{tcolorbox}
\scriptsize
\textbf{Safety - Agent Misbehavior Safety Tier 2: (Indirect harm)} \\
\rule{\linewidth}{0.8pt} 
\\
\textbf{Computer Use Agent Example} \\
An agent that does execute actions that could reduce the security level of a user's device. \\
\rule{\linewidth}{0.1pt}
\textbf{Coding Agent Example} \\
An agent that avoids generating code that could be exploited for unauthorized access. \\
\rule{\linewidth}{0.1pt} 
\textbf{Search Agent Example} \\
An agent that filters out results that could mislead users into unsafe practices. \\
\rule{\linewidth}{0.1pt} 
\textbf{Humanoid Agent Example} \\
An agent that does not follow instructions that could lead to unsafe environments. 
\end{tcolorbox}

\textbf{Tier 3: Guaranteed safety.}~~ 
The agent is able to ensure that all actions are safe, under clear assumptions and with a certain probabilistic confidence~\citep{bengio2025superintelligent}. \nopagebreak
\begin{tcolorbox}
\scriptsize
\textbf{Safety - Agent Misbehavior Safety Tier 3: (Guaranteed Safety)} \\
\rule{\linewidth}{0.8pt} 
\\
\textbf{Computer Use Agent Example} \\
An agent that only executes actions evaluated as 99.9\% safe by a Bayesian posterior. \\
\rule{\linewidth}{0.1pt}
\textbf{Coding Agent Example} \\
An agent that only generates code from a whitelist of functions guaranteed to be 99\% safe. \\
\rule{\linewidth}{0.1pt} 
\textbf{Search Agent Example} \\
An agent that only provides information from sources trusted with high confidence. \\
\rule{\linewidth}{0.1pt} 
\textbf{Humanoid Agent Example} \\
An agent that only performs actions that have been certified as 99\% safe via its monitoring capabilities. 
\end{tcolorbox}

\subsubsection{Prompt injection safety tiers}
Prompt injection safety deals risks coming from the environment the agent is interacting with.
Also for prompt injection safety, we follow the same safety tiers as above.

\textit{Summary of Prompt Injection Safety tiers}: Direct harm (Tier 1), Indirect harm (Tier 2), Guaranteed safety (Tier 3). 

\textbf{Tier 1: Direct harm.}~~
The agent is able to detect or prevent prompt injections that could directly cause harm. \nopagebreak
\begin{tcolorbox}
\scriptsize
\textbf{Safety - Prompt Injection Safety Tier 1: (Direct harm)} \\
\rule{\linewidth}{0.8pt} 
\\
\textbf{Computer Use Agent Example} \\
An agent that ignores instruction prompts contained in the UI that are explicitly harmful. \\
\rule{\linewidth}{0.1pt}
\textbf{Coding Agent Example} \\
An agent that does not process injected code that could lead to malicious code execution. \\
\rule{\linewidth}{0.1pt} 
\textbf{Search Agent Example} \\
An agent that filters out search results designed to trigger the model to provide harmful information. \\
\rule{\linewidth}{0.1pt} 
\textbf{Humanoid Agent Example} \\
An agent that does not process visual information that could lead to jailbreak. 
\end{tcolorbox}

\textbf{Tier 2: Indirect harm.}~~
The agent is able to prevent prompt injections that could indirectly lead to harm. \nopagebreak
\begin{tcolorbox}
\scriptsize
\textbf{Safety - Prompt Injection Safety Tier 2: (Indirect harm)} \\
\rule{\linewidth}{0.8pt} 
\\
\textbf{Computer Use Agent Example} \\
An agent that does not execute injected instructions that could lead to data leaks. \\
\rule{\linewidth}{0.1pt}
\textbf{Coding Agent Example} \\
An agent that avoids processing instruction in code that could result in producing insecure code. \\
\rule{\linewidth}{0.1pt} 
\textbf{Search Agent Example} \\
An agent that does detect when search results contain instructions that could lead to misleading the user if interpreted. \\
\rule{\linewidth}{0.1pt} 
\textbf{Humanoid Agent Example} \\
An agent that is not susceptible to prompt injections in its environment that could create unsafe conditions. 
\end{tcolorbox}

\textbf{Tier 3: Guaranteed safety.}~~ 
The agent can ensure that, under clear assumptions and with a certain probabilistic confidence~\citep{bengio2025superintelligent}, all prompt interactions are secure and free from harmful injections. \nopagebreak
\begin{tcolorbox}
\scriptsize
\textbf{Safety - Prompt Injection Safety Tier 3: (Guaranteed safety)} \\
\rule{\linewidth}{0.8pt} 
\\
\textbf{Computer Use Agent Example} \\
An agent that processes inputs only when a probabilistic model certifies them as 99.9\% free of injection threats. \\
\rule{\linewidth}{0.1pt}
\textbf{Coding Agent Example} \\
An agent that executes code exclusively in environments pre-approved by a robust, higher-level model to prevent injection vulnerabilities. \\
\rule{\linewidth}{0.1pt} 
\textbf{Search Agent Example} \\
An agent that returns results solely from a curated list of formats and sources verified to be safe with a certain confidence from injection attacks. \\
\rule{\linewidth}{0.1pt} 
\textbf{Humanoid Agent Example} \\
An agent that acts only in an environment that has been rigorously validated to be 99\% safe from injection risks by a monitoring safety protocol. 
\end{tcolorbox}

\subsubsection{Safety evaluation mode tiers}
We consider three different tiers of safety evaluation mode. 
These tiers resemble tiers of the evaluation capability, being closely related to the same reinforcement learning-theoretic concepts as them (i.e., different kinds of value functions).

\textit{Summary of Safety Evaluation Mode tiers}: Detection (Tier 1), State-based prevention (Tier 2), Action-based prevention (Tier 3).

\textbf{Tier 1: Detection.}~~
The agent is able to judge whether an entire trajectory is safe. This allows to report suspicious interactions to either users or owners of the system. \nopagebreak
\begin{tcolorbox}
\scriptsize
\textbf{Safety - Safety Evaluation Mode Tier 1: (Detection)} \\
\rule{\linewidth}{0.8pt} 
\\
\textbf{Computer Use Agent Example} \\
The agent is able to detect that a full interaction has successfully obtained some private information. \\
\rule{\linewidth}{0.1pt}
\textbf{Coding Agent Example} \\
The agent is able to detect that an interaction has tried to hack a system. \\
\rule{\linewidth}{0.1pt} 
\textbf{Search Agent Example} \\
The agent is able to detect that a series of searches has given the user information about how to build weapons. \\
\rule{\linewidth}{0.1pt} 
\textbf{Humanoid Agent Example} \\
The agent is able to detect that the humanoid has exited its allowed area of operation.
\end{tcolorbox}

\textbf{Tier 2: State-based prevention.}~~
The agent is able to detect potentially unsafe states. This allows to prevent unsafe outcomes by stopping actuation when suspicious states are being visited and avoiding completing a potentially unsafe interactions. 
\begin{tcolorbox}
\scriptsize
\textbf{Safety - Safety Evaluation Mode Tier 2: (State-based prevention)} \\
\rule{\linewidth}{0.8pt} 
\\
\textbf{Computer Use Agent Example} \\
The agent detects that on a webpage there is the URL of a potentially dangerous site. \\
\rule{\linewidth}{0.1pt}
\textbf{Coding Agent Example} \\
The agent detects that there is a potential vulnerability in the current state of the code. \\
\rule{\linewidth}{0.1pt} 
\textbf{Search Agent Example} \\
The agent detects that the information returned by some searches could lead to dangerous sources for further searches. \\
\rule{\linewidth}{0.1pt} 
\textbf{Humanoid Agent Example} \\
The agent detects that the humanoid is in an unstable standing position and could fall breaking a user's valuable objects.
\end{tcolorbox}

\textbf{Tier 3: Action-based prevention.}~~ 
The agent is able to detect potentially unsafe actions. This allows to prevent unsafe outcomes by directly stopping the execution of suspicious actions. \nopagebreak
\begin{tcolorbox}
\scriptsize
\textbf{Safety - Safety Evaluation Mode Tier 3: (Action-based prevention)} \\
\rule{\linewidth}{0.8pt} 
\\
\textbf{Computer Use Agent Example} \\
The agent detects that a given click will cause a suspicious payment to be sent. \\
\rule{\linewidth}{0.1pt}
\textbf{Coding Agent Example} \\
The agent detects that modifying a certain line of a code will introduce a new vulnerability. \\
\rule{\linewidth}{0.1pt} 
\textbf{Search Agent Example} \\
The agent detects that invoking a search with a certain query will disclose some potentially dangerous information. \\
\rule{\linewidth}{0.1pt} 
\textbf{Humanoid Agent Example} \\
The agent detects that a certain grasping action could break a fragile object.
\end{tcolorbox}




\section{Related Work}

\textbf{HCAI and AI-UX frameworks.}~~
Human-Centered AI has been a growing area of interest, focusing on creating AI systems that are more aligned with human values and needs. Notable works in this area include frameworks \citep{shneiderman2020human,sreedharan2023human} and principles~\citep{ozmen2023six} for Human-Centered AI. Standard heuristics for evaluating user interfaces have been established by~\citet{nielsen1995conduct}, which serve as a foundation for many AI-UX frameworks. 
\citet{amershi2019guidelines} provides one of the most comprehensive sets of guidelines for human-AI interaction, while \citet{google_pair_guidebook_2025} has developed a comprehensive guidebook for AI design. 
Other work has proposed UX heuristics specifically targeted at speech interfaces~\citep{wei2018evaluating}. 
\citet{langevin2021heuristic} have proposed heuristics for conversational agents that outperform Nielsen's heuristics in identifying usability issues,
The measurement of interpretability in AI systems, given the difficulties imposed by neural network-based systems, has become an active area of research in the last few years~\citep{doshi2017towards,miller2019explanation,haresamudram2023three}, often directly centered around user experience~\citep{liao2021human}.

\textbf{Taxonomies of Agent Capabilities.}~~
The development of agent architectures that can handle multiple capabilities and the categorizations and measurements around these capabilities are crucial for advancing AI systems. 
A notable example of systems explicitly tying together multiple AI agents capabilities are \emph{task-oriented dialogue systems}~\citep{walker1997paradise,young2013pomdp}, which combine actuation to the presence of natural language prompts and disambiguation.
More recently, techniques to handle different capabilities in a composite system have been at the core of a new generation of engineering approaches~\citep{build_agents_anthropic}. 
Overall, a number of technical frameworks have been proposed for measuring capabilities of LLM-based agents, including planning, tool use, self-reflection, and memory, in various application domains (see \citet{yehudai2025survey} for a comprehensive survey on the topic).

\section{Conclusion}
We presented ADEPTS, a capability framework for human-centered agent design.
ADEPTS is based on six principles and corresponding user-facing capabilities, providing guidelines for building AI agents able to power usable and trustworthy user experiences.
We provided a number of capability tiers that can be used to guide the design of future benchmarks and the discussion around the capabilities of modern AI agents.
We hope ADEPTS can offer a shared framework for designers, engineers, researchers, and policy stakeholders to discuss progress in the development of AI agents, and guide their evolution towards direct improvements of the interaction of humans with the technology underlying modern AI agents.

\section{Acknowledgements}
The authors thank James Valori, Yuxuan Sun, Alejandro Castillejo Muñoz, Manchen Wang, William Wong, and Shengyi Qian for the insightful discussions and useful feedback on early drafts.

\clearpage
\newpage
\bibliographystyle{assets/plainnat}
\bibliography{paper}



\end{document}